%% file: main.tex
\documentclass[letterpaper, 10pt, conference]{ieeeconf}  

\IEEEoverridecommandlockouts                              

\usepackage{amsmath} 

\usepackage{amssymb}  
\usepackage{color}
\usepackage{bm}
\usepackage{array}
\usepackage{cite}
\usepackage[utf8]{inputenc}
\usepackage{makecell}
\usepackage{xcolor}
\usepackage{siunitx}
\usepackage{multirow}
\usepackage[bottom]{footmisc}
\usepackage{graphicx}
\usepackage{url}
\usepackage[hidelinks]{hyperref}
\usepackage{verbatim}
\usepackage{url}
\usepackage[shortcuts]{extdash} 
\usepackage{microtype}
\usepackage{caption}
\usepackage{subcaption}
\usepackage{fancyhdr}
\fancypagestyle{withfooter}{
  
  \fancyfoot[C]{\footnotesize Accepted to the IEEE ICRA Workshop on Field Robotics 2024}
}

\newcommand{\BibTeX}{\rm B\kern-.05em{\sc i\kern-.025em b}\kern-.08em\TeX}

\title{\LARGE \bf \emph{AutoInspect}: Towards Long-Term Autonomous Industrial Inspection}

\author{Michal Staniaszek$^1$, Tobit Flatscher$^1$, Joseph Rowell$^{1}$, Hanlin Niu$^2$, Wenxing Liu$^2$, Yang You$^2$, \\Robert Skilton$^2$, Maurice Fallon$^1$, Nick Hawes$^1$%
\thanks{Initial work was supported by the Innovate UK AutoInspect grant (1004416). Further work was supported by the EPSRC Programme Grant “From Sensing to Collaboration” (EP/V000748/1), the UKAEA/EPSRC Fusion Grant (EP/W006839/1), and part of the work has been carried out within the framework of the EUROfusion Consortium, funded by the European Union via the Euratom Research and Training Programme (Grant Agreement No 101052200 — EUROfusion). Views and opinions expressed are those of the author(s) and do not necessarily reflect those of the European Union or the European Commission. Neither the European Union nor the European Commission can be held responsible for them.}%
  \thanks{$^{1}$Oxford Robotics Institute, University of
  Oxford, Oxford, UK. {\tt\footnotesize \{michal, tobit, joseph, mfallon, nickh\}@robots.ox.ac.uk}.}%
  \thanks{$^{2}$Remote Applications in Challenging Environments (RACE), United Kingdom Atomic Energy Authority, Culham, UK. {\tt\footnotesize \{hanlin.niu, wenxing.liu, yang.you, robert.skilton\}@ukaea.uk}.}
 } 
\begin{document}

\maketitle
\thispagestyle{withfooter}
\pagestyle{withfooter}

\input{chapters/1_abstract.tex}

\input{chapters/2_intro.tex}
\input{chapters/3_method.tex}
\input{chapters/4_experiments.tex}
\input{chapters/5_conclusion}

\bibliographystyle{IEEEtran}
\bibliography{IEEEAbrv,refs_short}

\end{document}

%% file: chapters/1_abstract.tex
\begin{abstract}
We give an overview of \emph{AutoInspect}, a ROS-based software system for robust and extensible mission-level autonomy. Over the past three years AutoInspect has been deployed in a variety of environments, including at a mine, a chemical plant, a mock oil rig, decommissioned nuclear power plants, and a fusion reactor for durations ranging from hours to weeks. The system combines robust mapping and localisation with graph-based autonomous navigation, mission execution, and scheduling to achieve a complete autonomous inspection system. The time from arrival at a new site to autonomous mission execution can be under an hour. It is deployed on a Boston Dynamics Spot robot using a custom sensing and compute payload called \emph{Frontier}. In this work we go into detail of the system's performance in two long-term deployments of 49 days at a robotics test facility, and 35 days at the Joint European Torus (JET) fusion reactor in Oxfordshire, UK.
\end{abstract}

%% file: chapters/2_intro.tex
\section{Introduction}
The increasing capabilities of commercially available robot platforms means it is possible to deploy robots in more challenging environments, both man-made and natural. Opening up new applications for mobile autonomy, these advances mean that research in planning, mapping, and inspection can be applied in industrial scenarios to demonstrate commercial relevance. This requires robust and flexible autonomy systems which can be used for a variety of applications.

\emph{AutoInspect} is such an autonomy system. It brings together our system for robot localisation and mapping (SLAM), called VILENS~\cite{wisth2023Vilens,ramezani2020Slam} with a graph-based topological autonomy system, to create a complete system for large-scale autonomous navigation and mission execution. The mapping system builds accurate 3D pointclouds of environments that the robot can then localise against. This allows the robot to navigate autonomously across multi-floor buildings, and ensures repeatability when carrying out potentially location-sensitive actions such as image capture. Our topological autonomy system abstracts the SLAM map by introducing a topological graph, allowing the robot to navigate through the environment autonomously with control over how each graph edge is traversed. It also provides operators with tools to build, edit, and visualise navigation graphs, create missions to execute sequences of tasks, and schedules to repeatedly perform missions at required times. In combination, this results in a robust and practical autonomy system.

The system is deployed in two stages. First, we build a map of the environment by teleoperating the robot around a facility while running our SLAM tool chain. During this stage a pointcloud and pose graph are generated in real-time. At the end of the mapping run the pose graph is converted into a topological map, then modified by operators to add inspection points or locations of charging docks. Robot actions, missions, and schedules are then configured according to operator requirements. In the second stage, the robot operates fully autonomously, localising within the pointcloud, and following the schedule or on-demand missions issued by operators.

\begin{figure}
    \centering
    \includegraphics[width=\columnwidth]{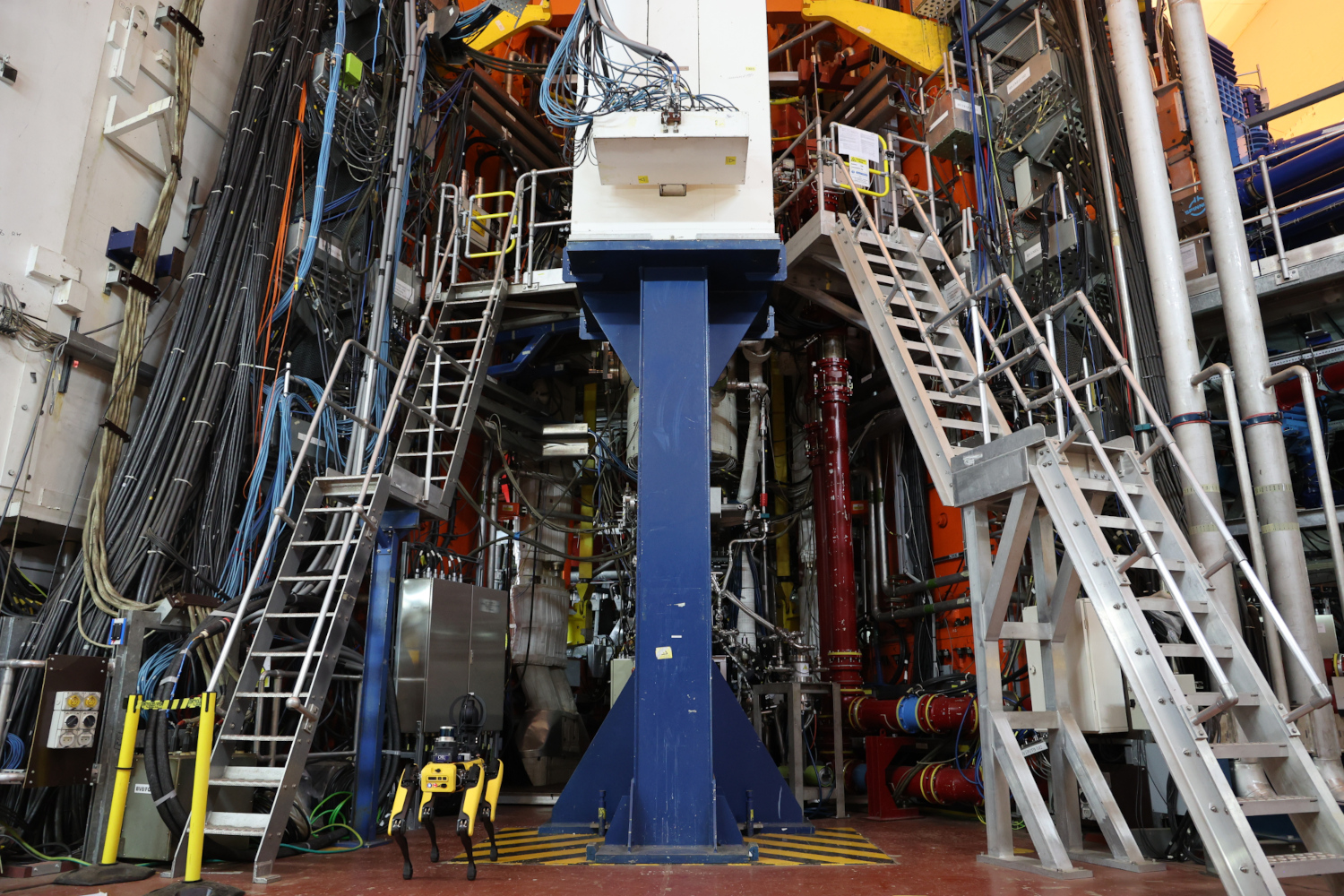}
    \caption{Spot measuring gamma radiation near the reactor during long-term deployment at the Joint European Torus.}
    \label{fig:jet-deployment}
\end{figure}

AutoInspect is currently integrated on the \emph{Frontier} device, shown in Figure~\ref{fig:frontier_and_spot}, mounted on the Boston Dynamics Spot, but the system can be used on any platform that meets integration requirements. Figure~\ref{fig:autoinspect_abstract} shows an overview of the software system as a whole, which is implemented using ROS Noetic \cite{quigley2009Ros}. It bears some resemblance to platform-specific industrial inspection solutions implemented for the ANYbotics ANYmal~\cite{bellicoso2018Anymal, gehring2021Anymal, anymal-capability} or Spot\cite{bd-orbit}, which try to allow for flexible payload integration for different applications. In contrast to many inspection systems~\cite{lattanzi2017ReviewRoboticInfrastructure, loupos2018AutonomousRoboticSystem, rocha2021ROSIRoboticSystem}, we avoid close integration with specific robots or objectives. The graph-based navigation system provides standard interfaces to integrate robot-specific navigation commands, and other interfaces can connect mission execution to modules which provide the robot with the ability to perform actions.

Development of the \emph{AutoInspect} system is based on the following requirements:

\begin{enumerate}
    \item Provide standard interfaces to facilitate integration of different robot capabilities 
    \item Capable of running autonomously for long periods of time with minimal human intervention    
    \item Easy for researchers and industrial partners to use 
    \item Adaptability to different robots and hardware configurations
\end{enumerate}

The system in its current state fulfills the first two requirements. We are in the process of testing the ease of use requirement in deployments with partners, and have developed documentation, user interfaces and system workflows which attempt to minimise the technical expertise needed to operate the system. The final requirement is untested. While we have used system components on a variety of robots including another quadruped, a drone, and service robots, we are now working to deploy the full system on an outdoor UGV.

Deployments thus far of the combined system have been focused on the Spot. During system development we performed trials in a chemical plant, a decommissioned nuclear power plant, a mine, a mock oil rig, and around partially demolished buildings at an outdoor training ground for firefighters. These initial trials of the system allowed us to assess robustness, add missing features, and improve end-user workflow as well as benefiting from interactions with industrial partners.

The following sections describe the system as implemented on the Boston Dynamics Spot using our Frontier device, with Section~\ref{sec:hardware} covering the hardware, Section~\ref{sec:mapping} the mapping and localisation, and Section~\ref{sec:autonomy} the graph-based autonomy system. Finally, Section~\ref{sec:deployment} summarises two long-term deployments of 49 days at a robotics test facility and 35 days at the JET fusion reactor in Oxfordshire, UK.

%% file: chapters/3_method.tex
\begin{figure}
    \centering
    \includegraphics[width=0.49\columnwidth]{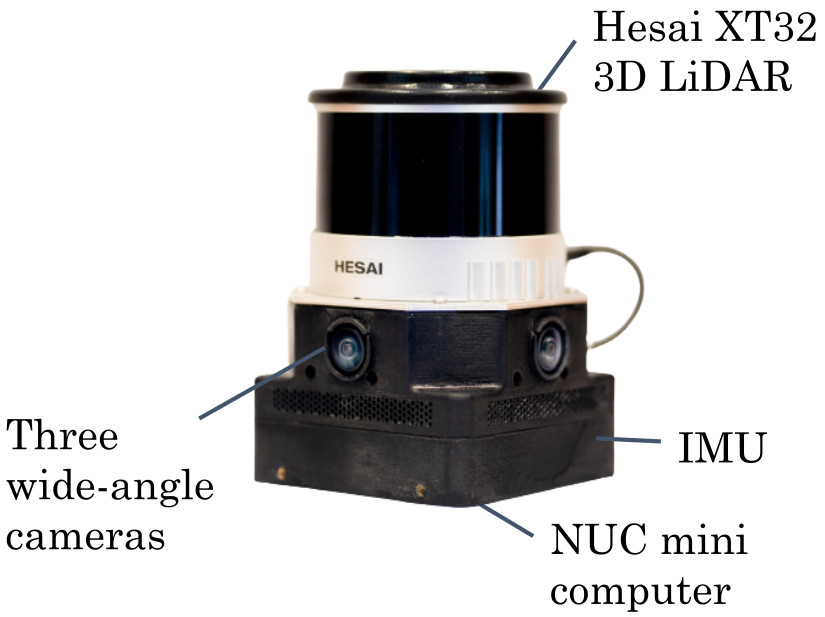}
    \includegraphics[width=0.49\columnwidth]{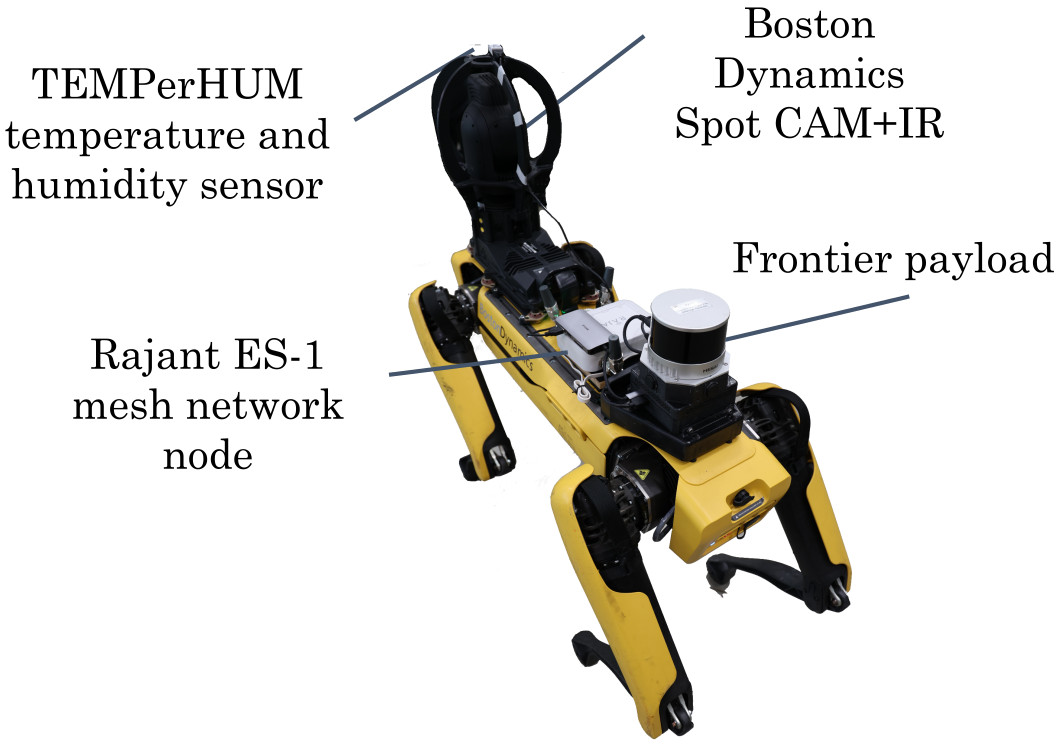}
    \caption{\emph{Left}: Our autonomy payload, \emph{Frontier}, consisting of a Hesai LiDAR, three Sevensense Alphasense cameras, and an IMU. \emph{Right}: Boston Dynamics Spot with the payloads used for deployments at UKAEA. The Frontier runs the entire AutoInspect software stack and sensor drivers.}
    \label{fig:frontier_and_spot}
\end{figure}

\section{Hardware}
\label{sec:hardware}

We chose to integrate AutoInspect on Spot primarily because it has reliable collision avoidance and stair climbing capabilities, which fit well with our intended application areas, and we can use its autonomous docking capability to facilitate long-term deployments.

The key hardware component of our system is our \emph{Frontier} autonomy payload (Figure~\ref{fig:frontier_and_spot}). It combines a NUC mini-PC with i7-1165G7@2.8 GHz and 32 GB of RAM, a Hesai XT-32 LiDAR, three time-synchronised fish-eye Sevensense Alphasense cameras, and a Bosch BMI085 IMU in a compact 3D printed case, with a combined weight of 1.5~kg. Camera intrinsics and extrinsics are calibrated with Kalibr~\cite{rehder2016Kalibr} while the camera-to-LiDAR extrinsics are calibrated according to~\cite{fu2023LidarCalibration}. As all sensors are mounted rigidly, we can move the Frontier between robots without having to recalibrate the system. The Frontier is attached to Spot with a mounting plate which connects it to the robot's network and power supply through a payload port.

\begin{figure}
    \centering
    \includegraphics[width=\columnwidth]{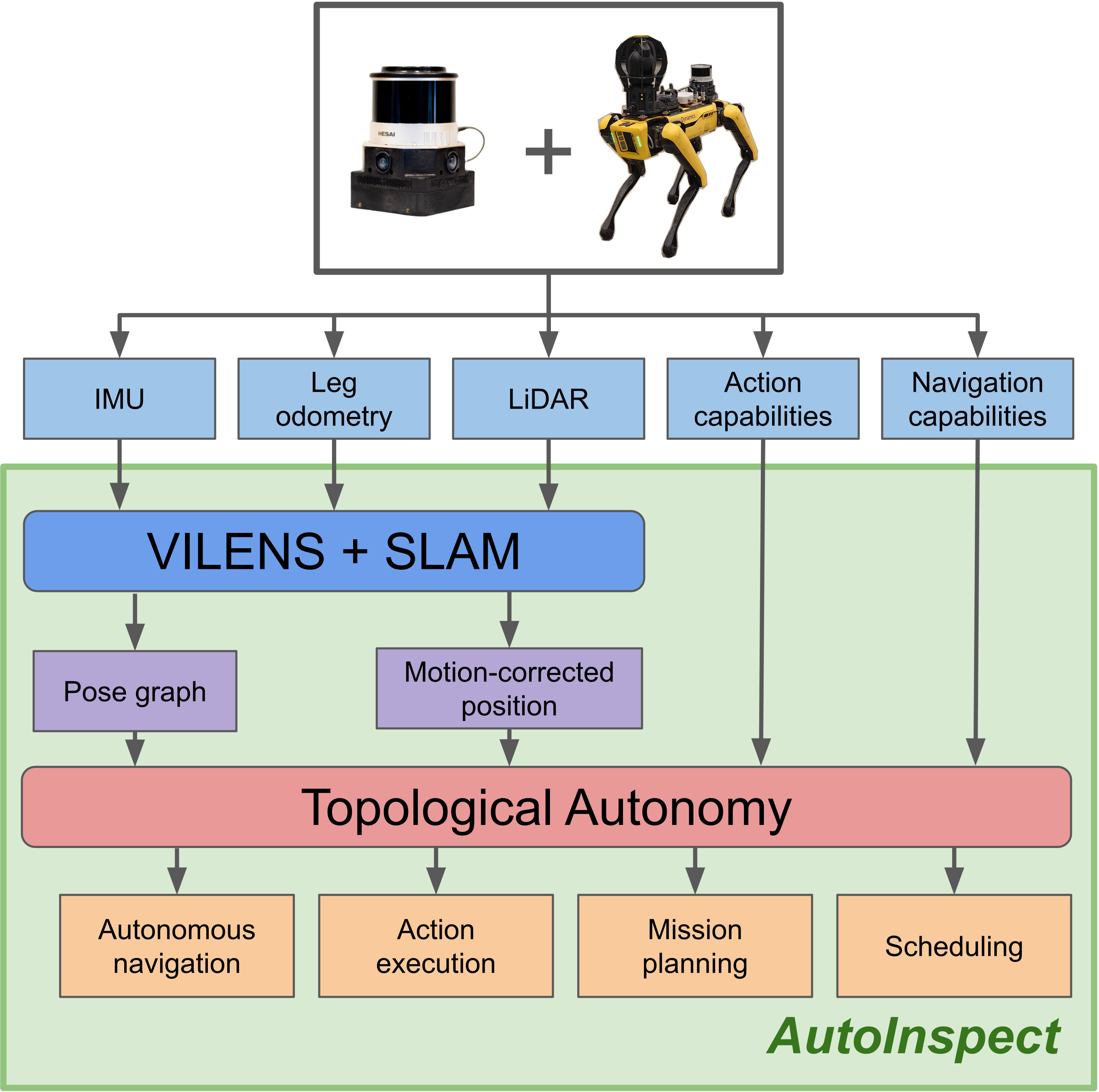}
    \caption{Overview of the AutoInspect system on Spot, using the Frontier payload: IMU measurements and LiDAR data from Frontier, as well as leg odometry from Spot are used by VILENS to generate a continuous pose estimate of the robot and a SLAM pose graph. These inputs, was well as Spot's action and navigation capabilities, are used by the topological autonomy system to provide autonomous navigation, action execution, mission planning, and scheduling capabilities to the operator.}
    \label{fig:autoinspect_abstract}
\end{figure}

Operators monitor the system using an external computer while it is running. To maintain network connectivity we use an industrial-grade mesh network. One Rajant {ES\=/1} network node is mounted on the robot, and another is connected to the operator computer. Other nodes are set up in the environment to cover the operational area of the robot. Network connectivity across an entire site is only necessary if constant monitoring is required. The AutoInspect system is self-contained on the Frontier and does not require a network connection except to receive operator commands.

In addition to the Frontier and the Rajant mesh node, we also use the manufacturer-supported SpotCAM+IR which has a pan-tilt-zoom unit with thermal and visual cameras. Other sensors we use in the experiments in Section~\ref{sec:deployment} include a temperature and humidity sensor and a Kromek Sigma 50 Gamma ray detector, both connected to the Frontier device.

\section{Mapping, Localisation, and Change Detection}
\label{sec:mapping}

The mapping and localisation system uses the VILENS odometry~\cite{wisth2023Vilens} and SLAM systems~\cite{ramezani2020Slam}, which fuse IMU, leg odometry, and LiDAR to provide robust and accurate odometry. (In the case of Spot, the leg odometry is provided by the Boston Dynamics API). Because of the dynamic, jerky motion of the robot, it is important to correct the motion of the LiDAR during each scanning sweep to achieve the most accurate maps. Previous applications of the VILENS system include  construction~\cite{zhang2022Hilti}, forestry \cite{proudman2022Forestry}, and aerial inspection with a drone~\cite{border2023Osprey}. The change detection system is a more recent addition which we use to infer 3D object-level change between successive missions.

\subsection{Initial SLAM Mapping}
\begin{figure}
    \centering
    \includegraphics[width=\columnwidth]{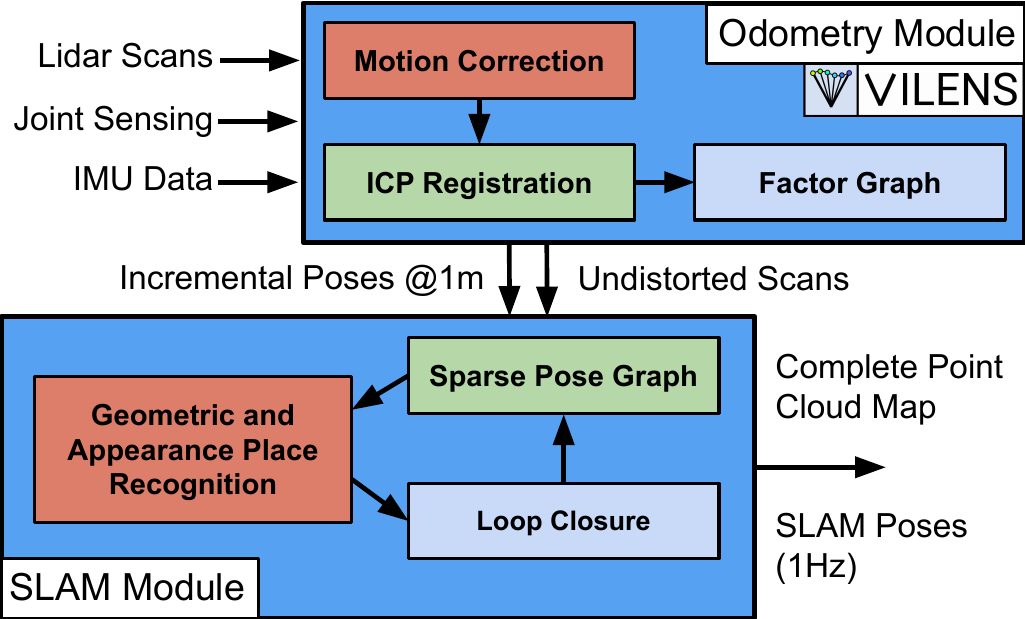}
    \caption{Overview of the VILENS odometry and VILENS SLAM system. On Spot, AutoInspect uses leg odometry as an additional input to the odometry module.}
    \label{fig:vilens_slam}
\end{figure}

To operate in a environment autonomously, our system localises in a prior map of the environment. The map can be generated from existing 3D LiDAR scans taken with a terrestrial LiDAR scanner but we typically create a map from scratch with our SLAM system. 
In the latter case odometry is provided by VILENS~\cite{wisth2023Vilens}, which can be configured to use different combinations of odometry sources such as IMU, visual feature tracking, LiDAR ICP registration, and leg odometry, and produces motion-corrected LiDAR scans. These corrected scans are passed to VILENS SLAM~\cite{ramezani2020Slam} to perform pose graph optimization using the iSAM2 solver~\cite{Dellaert2017}. Proposals for loop closures can come from geometric constraints as well as place recognition based on the ScanContext descriptor~\cite{kim2018ScanContext}. An overview of the system is shown in Figure~\ref{fig:vilens_slam}. The output of the mapping step is a pose graph with corresponding individual pointclouds, as well as a global map in which all individual clouds have been registered in a global reference frame.

\subsection{Subsequent Localisation in Prior Map}
To re-localize in a global prior map we use ICP which requires an accurate initial pose estimate to be provided by the operator or by initialising the robot in a known location. The pose estimate is then iteratively updated using the leg odometry motion prior and ICP to the prior map at 2Hz.

An alternative approach is to localize in a prior map made up of the individual pose-graph pointclouds using place recognition to determine the pose guess at each iteration. This approach does not require an initial pose estimate and can be particularly effective in very large environments where performing ICP to a single large pointcloud map would be computationally prohibitive.

\subsection{3D Change Detection}
AutoInspect incorporates LiSTA~\cite{rowell2024lista} which is a 3D change detection system. It uses volumetric differencing to detect object-level changes between pointclouds acquired across different missions. During each mission, a set of local pointclouds are acquired and converted into octrees. Octrees corresponding to the same spatial area, but from different missions, are then compared to generate a set of \textit{difference octrees} which are projected back into the original pointcloud. The pipeline includes ground filtering via RANSAC, Moving Least Squares smoothing, and morphological opening to obtain a set of pointcloud clusters.

The pointcloud clusters for each discovered object are then segmented out using Euclidean clustering. Finally, inter-mission correspondences of objects are determined through K-means clustering of SE(3) invariant descriptors assigned to each object using a learning-based 3D pointcloud descriptor~\cite{ZhangInstaLoc:Learning}. This provides actionable insights to operators in the form of pointclouds of object-level changes to the environment between missions. An application of the system is shown in Fig. \ref{fig:change-detection}

\section{Autonomy}
\label{sec:autonomy}

The second major component of the system is the autonomy component.  The autonomy system is made up of a topological map, topological navigation, mission planning and scheduling module as well as a user interface.

\begin{figure}
    \centering
    \includegraphics[width=\columnwidth]{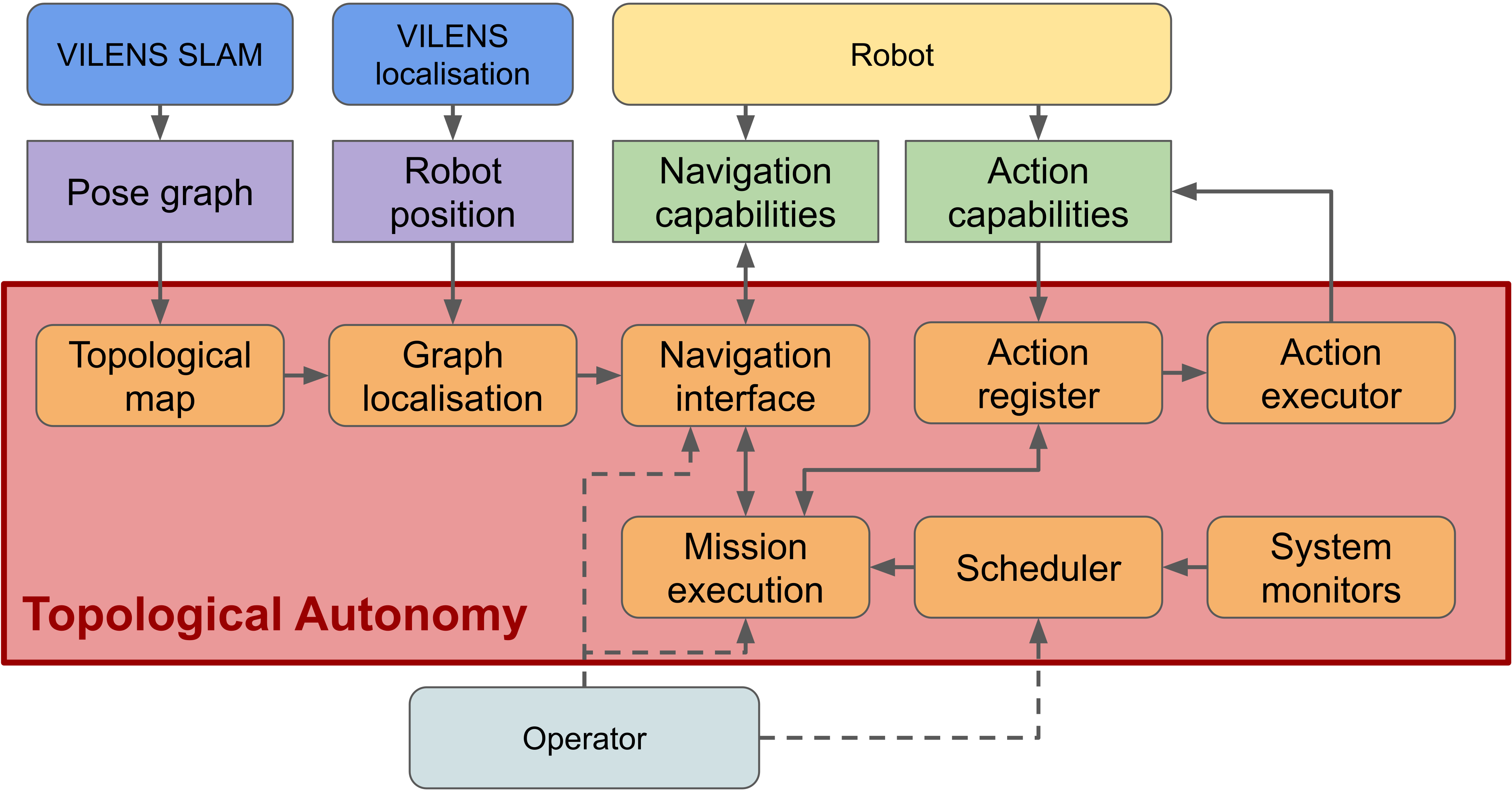}
    \caption{Overview of topological autonomy. The \texttt{Topological map} is constructed based on the VILENS SLAM pose graph, and the robot is localised on the map based on the 3D pose provided by localisation. The robot's action capabilities are registered with the \texttt{Action register}, which can call the \texttt{Action executor} to execute them. The robot's navigation capabilities are integrated into the \texttt{Navigation interface}. \texttt{Mission execution} navigates and executes tasks based on mission specifications. The \texttt{Scheduler} can execute missions on a schedule, subject to interrupts from \texttt{System monitors}. The operator typically controls the robot by scheduling or executing missions, or sending it to specific locations.}
    \label{fig:topological_autonomy}
\end{figure}

Topological autonomy is a modular system for autonomous navigation and mission execution which has at its core a topological map representation. In combination with the mapping and localisation system, this allows the robot to act in its environment without the need for continuous monitoring or input from operators. Topological map representations are good at representing large physical spaces~\cite{kuipers1978ModelingSpatialKnowledge}, and have been used extensively on mobile robots as a navigation abstraction since their introduction in this context by Brooks~\cite{brooks1985VisualMapMaking}. The topological map is well suited for incorporating domain knowledge during deployments and can be used to adapt the system's behaviour. It is also a natural representation for advanced planning and resource allocation algorithms, which is important for our research. Topological maps have been used as a basis for planning and navigation in field deployments in offices~\cite{kunze2012SearchingObjectsLargescale, mudrova2015IntegratedControlFramework, hawes2017STRANDSProjectLongTerm} and agriculture~\cite{das2023UnifiedTopologicalRepresentation}, as well as theoretical works~\cite{lacerda2019ProbabilisticPlanningFormal}. In addition to its technical benefits, it is also an intuitive representation for end users, providing a clear visual representation of the structure of the robot's operational area, and the ability to name locations in the environment is useful when communicating about missions. 

Topological maps, can be constructed in many ways, such as from aerial images~\cite{li2019TopologicalMapExtraction} and 2D or 3D maps~\cite{thrun1996IntegratingGridBasedTopological, zivkovic2006HierarchicalMapBuilding, blochliger2018TopomapTopologicalMapping}. We use a hybrid approach, automatically generating an initial map based on the SLAM pose graph from global map construction, then manually tailor it to the environment with our GUI tools.

Integrating a new robot with the autonomy system requires implementation of a localiser for the graph, an edge traversal interface, and defining its action capabilities with ROS actionlob goals. For example, with Spot, we use a localiser which monitors the TF frame of the robot's position in the global frame, and set up \texttt{actionlib} goals which specify positions for the Spot CAM+IR PTZ to take images at different locations. This gives us a structure which facilitates the core task of mission planning and scheduling, while being adaptable to different methods of localisation and navigation, and the varied action capabilities of different robots or sensors.

We have deployed topological autonomy on the Clearpath Husky, Toyota HSR, Hello Robot Stretch, and Boston Dynamics Spot robotic platforms, each of which are integrated differently. Our work on Husky uses visual teach and repeat for navigation, with no global map, and was used to autonomously monitor biology experiments in an outdoor environment~\cite{gadd2024watching}. The service robots HSR and Stretch share an implementation which integrates with the AMCL and the \texttt{move base} packages of the ROS navigation stack \cite{eppstein2010Navigation} and are used to conduct research in mobile manipulation.

\subsection{Topological Map}

\begin{figure}
    \centering
    \includegraphics[width=\columnwidth]{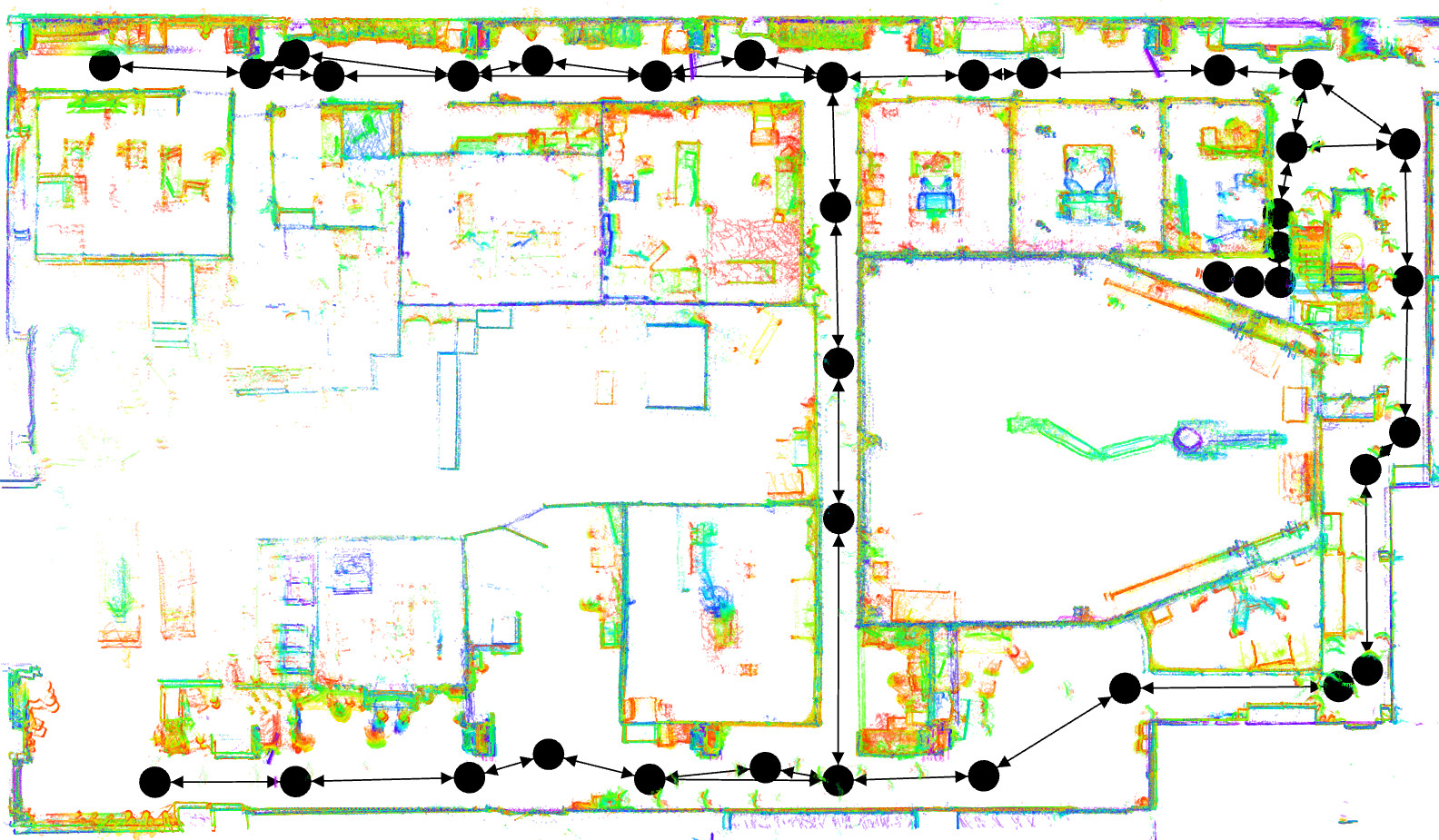}
    \caption{Top-down view of RACE B1 pointcloud with overlaid topological map. Points coloured by z coordinate height, red is lowest.}
    \label{fig:b1_topmap}
\end{figure}

The topological map is the core of the autonomy system, an example of which is shown in Figure~\ref{fig:b1_topmap}. The map is a graph made up of nodes and edges. An edge connecting two nodes indicates that the robot can move between those two locations. This is a simple yet powerful representation, as it allows us to annotate nodes and edges with domain-specific information for deployments, and provides a structured way to abstract the environment for input to planning algorithms, for example~\cite{lacerda2019ProbabilisticPlanningFormal}. The topology contains only an implicit spatial representation based on the nodes and edges. For most deployments, we augment the nodes of the map with 3D positions indicating their location in the global reference frame. Our tools allow the topological map to be edited while the system is running with no interruption to autonomy.

\subsection{Topological Navigation}
The topological map is used as the basis for our navigation framework. To send the robot to a node in the graph, the shortest path through the map is computed based on costs associated with each edge. The path is executed by traversing each individual edge in sequence. If an edge cannot be traversed, for example when the path is blocked by an obstacle, the system autonomously reroutes by temporarily deactivating the offending edge and recomputing the shortest path.

These components are linked to the navigation capabilities of specific robots through interfaces which translate the requested edge traversals to commands for the robot's navigation system. For example, to control Spot we use the \texttt{spot\_ros} driver~\cite{spot-ros}, which connects to the robot's API. The 3D position of the target node is retrieved and sent as a goal to the driver. Associating different types of traversal actions with edges allows us to control how the robot behaves when travelling between nodes, with no further human input necessary during navigation once the association is made. For example, custom actions may be required for moving through doors or up stairs. Paths through the map can involve multiple different types of edge traversals. Furthermore, this approach can also be used to perform other actions while traversing an edge, such as imaging a section of pipe.

\begin{figure}
    \includegraphics[width=\columnwidth]{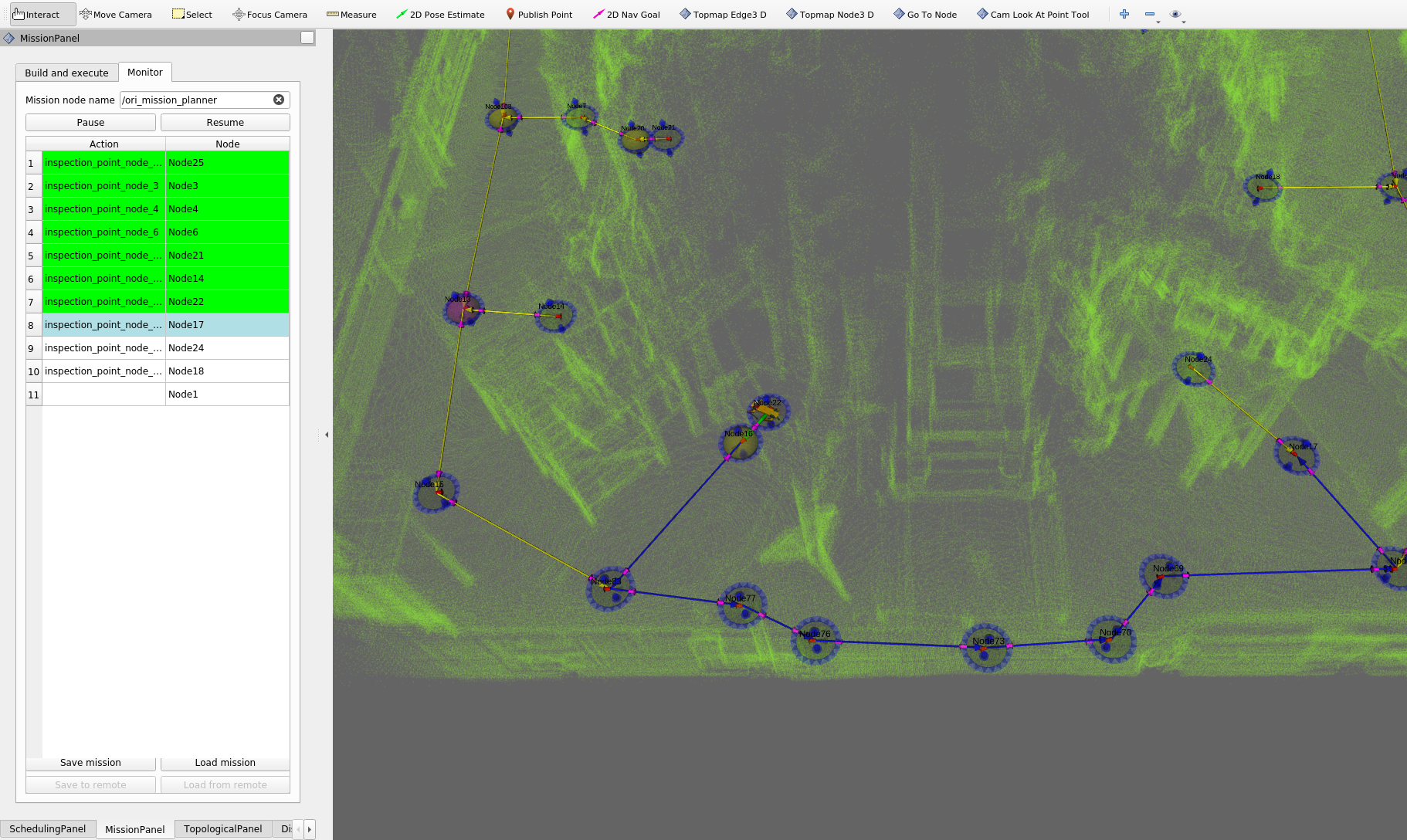}
    \caption{AutoInspect operating in the torus hall of the JET fusion reactor. The RViz interface shows the global map pointcloud and navigation graph. The mission monitor tracks progress of the current mission. Yellow arrows on edges indicate the global navigation policy, and blue arrows the expected path to the robot's next task location.}
    \label{fig:autoinspect_rviz}
\end{figure}

\subsection{Mission Execution and Scheduling}

The mission execution component links topological navigation and task execution. A task is defined as performing an action at a specific location. A mission consists of an arbitrary number of tasks and is usually specified by human operators, but can also be constructed programmatically. Missions are executed sequentially, navigating to the location of the next task, executing the action associated with the task, and repeating until all tasks are complete. Constructing missions manually while minimising total distance travelled can be challenging for humans, so the system can automatically minimise the path cost by reordering missions according to a solution to a standard travelling salesperson problem. Actions in a mission are associated with pre-defined ROS action servers implemented to execute the specific task. In addition to a set of predefined actions, users can dynamically register custom actions to fit their needs by providing an action server and a parametrised message to send to the action server when the action is executed. For example, in our deployments we make use of custom actions for controlling the Spot CAM+IR, and taking radiation and temperature measurements. Tasks in the mission may fail either because the task location cannot be reached, or the action itself fails. When this happens, we can configure the system to either attempt to complete remaining tasks in the mission, or abort.

The final layer of the autonomy system is the scheduler, which builds on the mission execution system and allows missions to repeat on a fixed schedule (e.g. every hour, every day at 09:00, every Tuesday at 16:00) or schedule a mission for execution once at a specific time. The scheduler includes monitors which, when specified conditions are fulfilled, can request execution of a mission, or disable execution of missions. For example, when the robot's battery is low, the battery monitor prevents any scheduled or user-requested mission from being executed, and immediately executes a mission which sends the robot to the nearest charging station.

\subsection{User Interface}

\begin{figure}
    \centering
    \includegraphics[width=0.3\columnwidth]{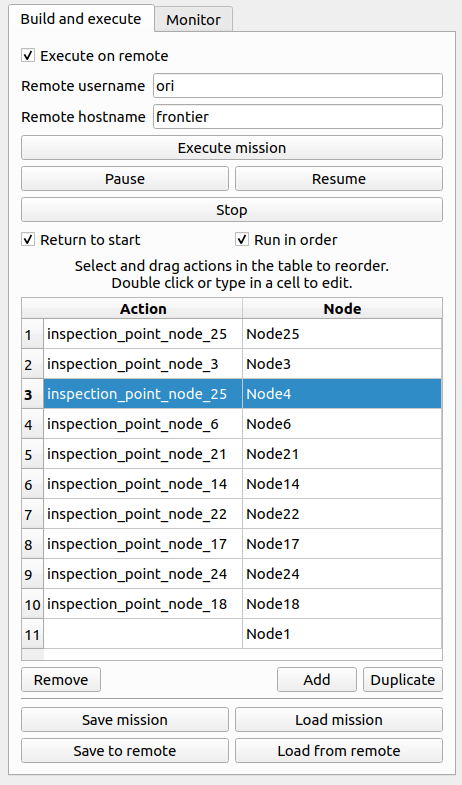}
    \includegraphics[width=0.3\columnwidth]{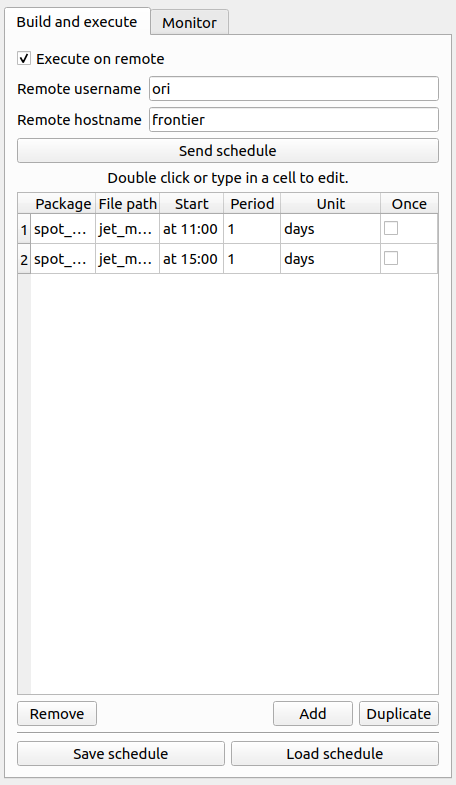}
    \includegraphics[width=0.3\columnwidth]{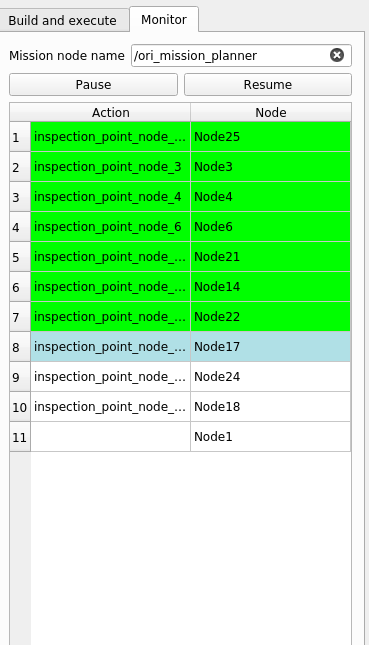}
        \caption{Examples of autonomy interfaces. \emph{Left}: mission construction, with action and node specification, remote and local saving, loading, and execution. Ordering of missions can be changed by dragging and dropping tasks. \emph{Centre}: schedule construction, with specification of predefined mission files and their repeat frequency and period. \emph{Right}: mission monitoring, where row colour indicates status of a task. Green tasks have been successfully completed, and light blue indicates the robot is navigating to the task location.}
    \label{fig:topological_ui}
\end{figure}

To provide a convenient and quick way for operators to interact with the system, we provide a graphical user interface implemented in the ROS RViz tool. Operators can modify topological maps in the RViz window, which allows repositioning, addition, or deletion of nodes and edges. RViz can also display the global 3D map which allows the topological map to be modified while taking into account features of the environment. The mission and scheduling panels allow operators to define, save, load, execute, or interrupt missions and schedules, and to monitor their status, as shown in Figures~\ref{fig:autoinspect_rviz}~and~\ref{fig:topological_ui}.

%% file: chapters/4_experiments.tex

\section{Deployments}
\label{sec:deployment}

In this section we will describe two long-term deployments of the system. Both took place at UKAEA's Culham Science Park in Oxfordshire UK, in collaboration with UKAEA's centre for Remote Applications in Challenging Environments (RACE). The first deployment was at the B1 robotics test facility, and the second in the torus hall of the JET fusion reactor, shown in Figure~\ref{fig:jet-deployment}. After initial startup of the hardware and software at the beginning of the deployment, we tracked interventions, which we define as any modification of the system's state by a person. \textit{Minor interventions} are normal usage of the system, such as modifying the topological map, running additional missions, or software failures which do not affect the system's functionality, such as logging. As such, we do not count minor interventions as interrupting the system. \textit{Serious interventions} are any modification during operation, or restarting small parts of the system. \textit{Fatal interventions} are restarts of the entire software stack or the hardware.

\subsection{RACE B1 Test Facility}
\begin{figure}
    \centering
    \includegraphics[width=\columnwidth]{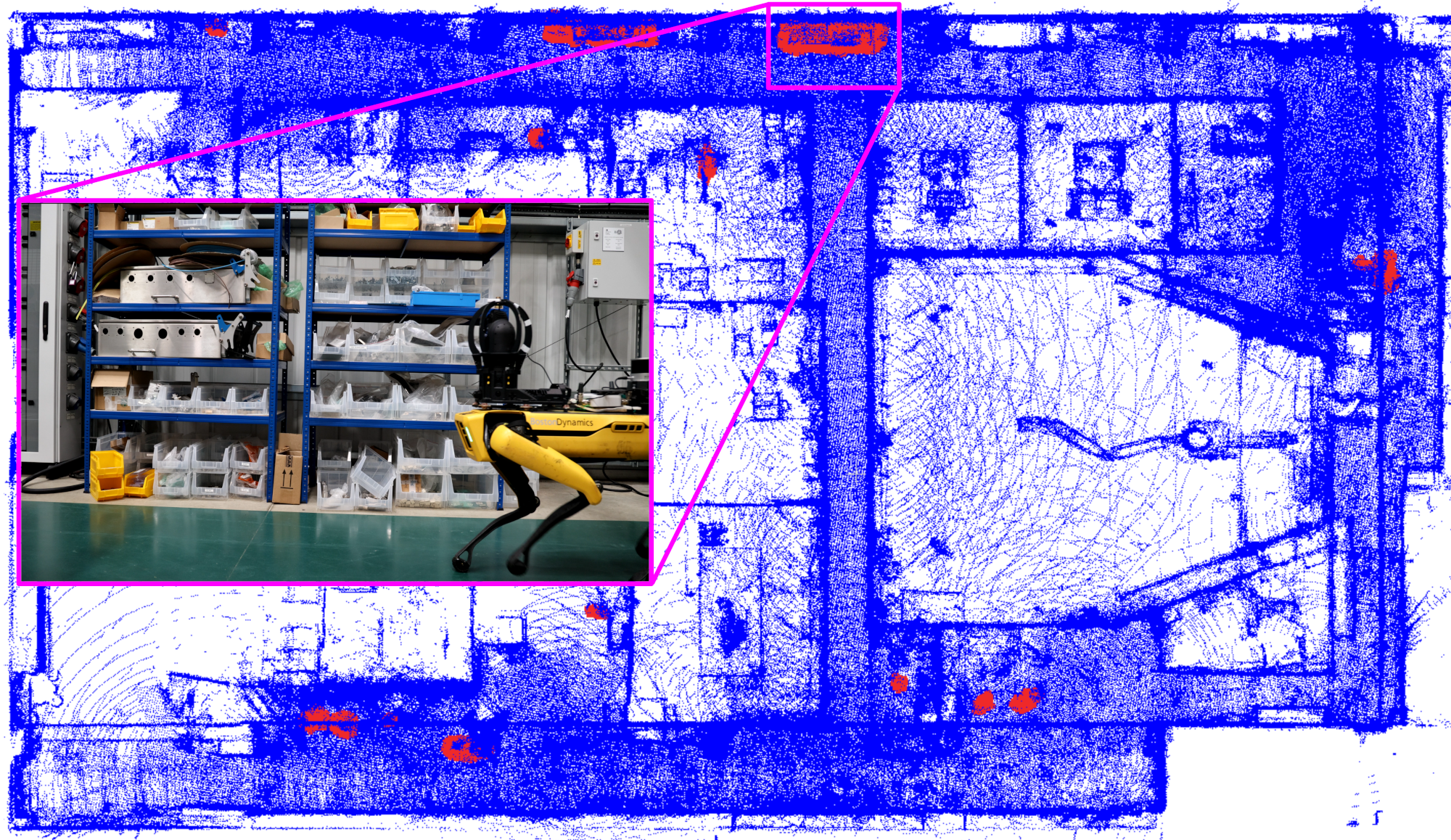}
    \caption{Change detection results from RACE deployment, showing intentionally induced changes between two missions in red, static cloud in blue. Moved shelving was unintentional.}
    \label{fig:change-detection}
\end{figure}
The robot was deployed at B1 research facility for total of 49 consecutive days from 18th July to 5th September 2023.  The topological map, shown in Figure~\ref{fig:b1_topmap}, consisted of 37 nodes, of which 9 were inspection points. At each inspection point, the robot captured temperature and humidity readings, as well as visual or thermal images. The SLAM system was run in a passive mode in order to capture new pointcloud maps for the offline change detection system. The robot performed a scheduled mission at 11:00 and 15:00 each weekday, with an average duration of 11 minutes 30 seconds. Including unscheduled missions for visitors, it completed 84 missions, during which 673 of 730 inspection actions were successfully performed. The robot walked 13.6~km over the 13 hours in which it was actively performing missions. Figure~\ref{fig:change-detection} gives an example of intentionally induced changes detected during the deployment, showcasing this important capability which provides operators with actionable insights.

We logged 25 interventions over the course of the deployment. The longest period without any serious or fatal interventions was 14 days. Twelve interventions were minor. The nine serious interventions included manually re-localising the robot (3 times), and restarting navigation (3), the camera driver (2), or localization (1). Four fatal interventions were restarting the entire software stack (3) or rebooting the Frontier payload (1). Not counting minor interventions, the mean time between interventions (MTBI) was 78 hours. In practice, interventions tended to be clustered together, likely due to configuration or hardware issues causing repeated failures. All of the interventions provided us with valuable information about how the system is used in practice, and how to improve its robustness. In this deployment, several interventions were caused by a bug in the localisation system which only manifested after long periods of time because of CPU throttling due to temperature build, which we may not have discovered without a long-term deployment.

\subsection{Joint European Torus (JET)}
\begin{figure}
    \centering
    \includegraphics[width=\columnwidth]{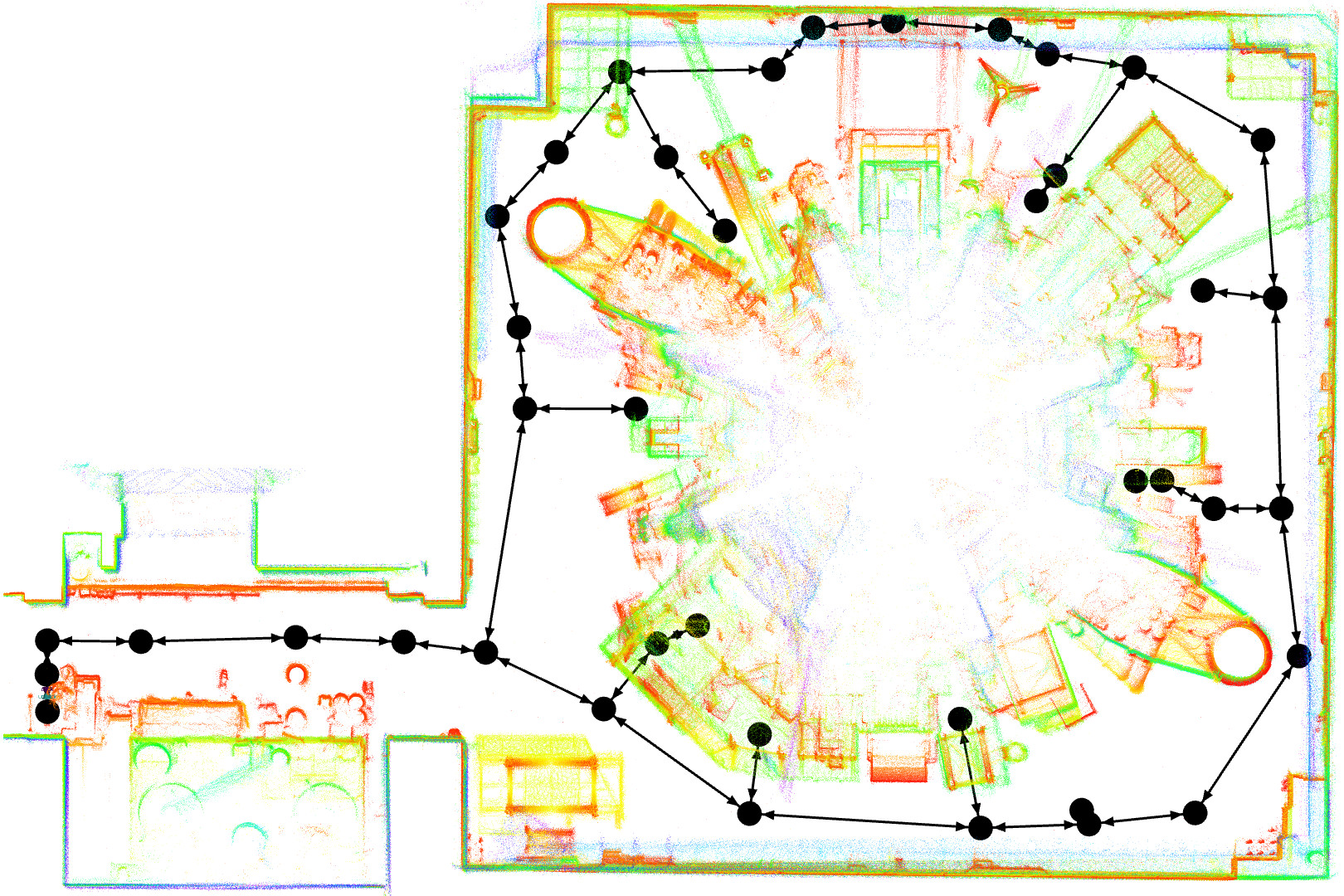}
    \caption{Top-down view of JET pointcloud with overlaid topological map. Points coloured by z coordinate height, red is lowest.}
    \label{fig:jet_topmap}
\end{figure}

\begin{figure}
    \centering
    \includegraphics[width=\columnwidth]{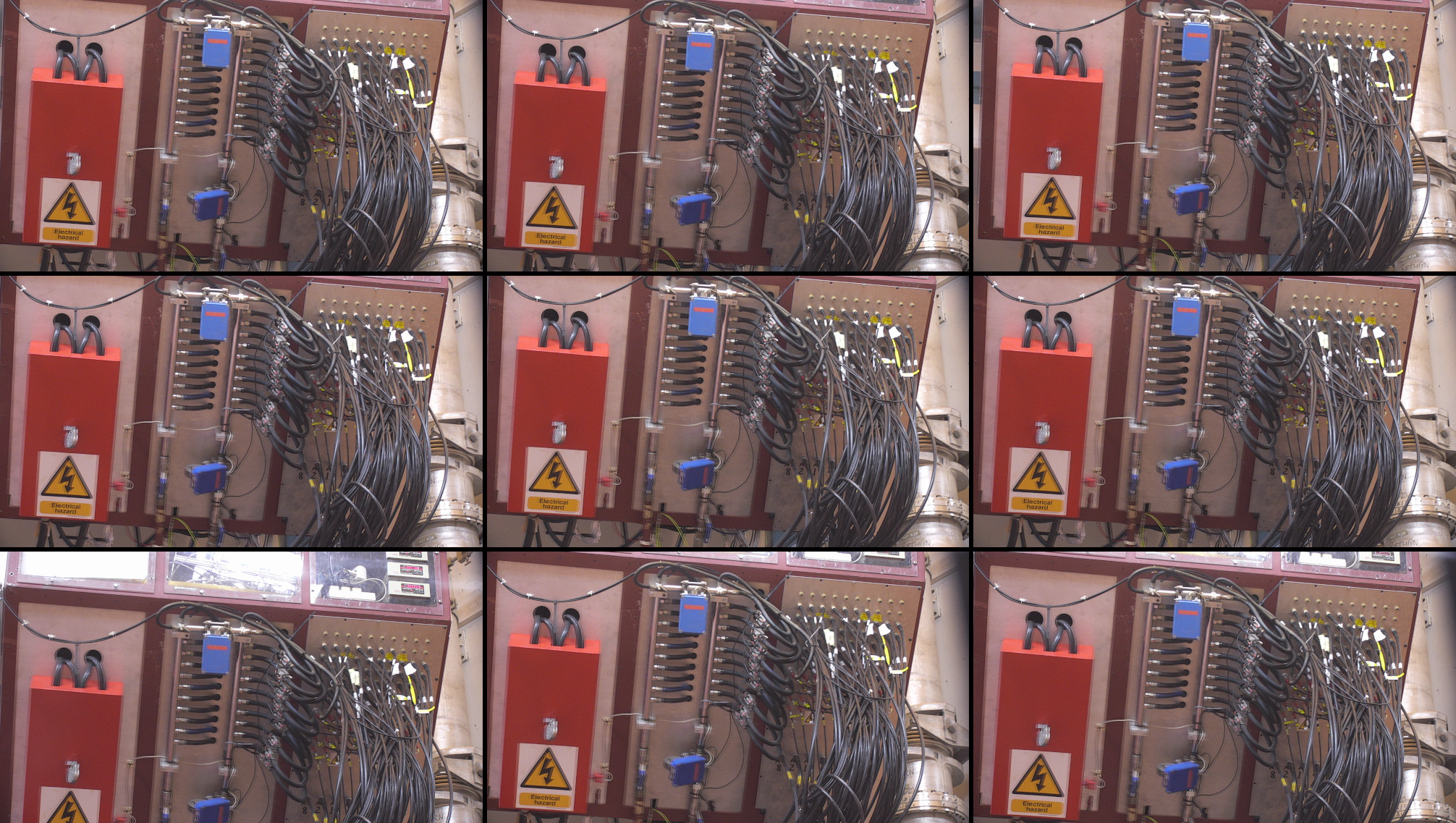}
    \caption{Inspection images of an electronics patch board taken with the Spot CAM+IR during the JET trial of object about 1.5m wide from a distance of approximately 5m. These nine images were randomly sampled from 60 taken at the same location. Their similarity indicates localisation robustness.}
    \label{fig:jet-images}
\end{figure}

Deployment in the torus hall of the JET fusion reactor ran for 35 consecutive days from 21st February to 27th March 2024, with an almost identical robot setup as for the RACE B1 trial. The topological map, shown in Figure~\ref{fig:jet_topmap}, had 41 nodes including 7 inspection points. Missions ran at 11:00 and 15:00, this time every day instead of just weekdays. In addition to the temperature and humidity sensors, we integrated a Kromek Sigma 50 gamma ray detector and created an action to record spectrum data for 1 minute. The average mission duration was 20 minutes, and the robot walked 15km over the 19 hours 30 minutes of its 81 missions. Figure~\ref{fig:jet-images} shows a sample of images taken from the same location across the deployment, which all appear very similar, showing localisation robustness and repeatability important for long-term deployments.

In this deployment, there were 16 interventions, which excluding minor interventions is an MTBI of 140 hours. The longest period without serious or fatal interventions was 15 days. 10 interventions were minor, 1 serious, and 5 fatal. The fatal interventions were caused by a user-facing process manager intermittently causing all ROS nodes to shut down when operators reconnected to the system. We noticed this bug at JET because manual adjustments of the schedule were more frequent due to construction and maintenance activities. We suspect this was the cause of 3 fatal interventions at B1, which we were unable to identify at the time. None of the interventions were caused by the core systems, indicating an increase in reliability compared to the B1 deployment.

%% file: chapters/5_conclusion.tex
\section{Conclusion}
We have presented an overview of AutoInspect, our mission-level mapping and autonomy system. We described two long-term deployments of the system in industrial inspection scenarios, during which the robot operated autonomously for 2 weeks without interruption, demonstrating its robustness.

The primary aim of future work is continued improvement in robustness of the system by stress-testing with increased up-time. In the near term, we plan to extend the system to multi-robot applications with a fleet of Clearpath Jackal UGVs, and continue to explore applications for autonomous inspection robots in other large-scale industrial sites. We will use the data from the initial deployments to design autonomous inspection routines and sensor suites to gather actionable scientific and operational data, and to demonstrate that autonomous robots can provide tangible benefits to operators.